\documentclass[conference]{IEEEtran}
\IEEEoverridecommandlockouts
\usepackage{amsmath,amssymb,amsfonts}
\usepackage{algorithmic}
\usepackage{graphicx}
\graphicspath{{figures/}}
\usepackage{textcomp}
\usepackage{xcolor}
\usepackage{subcaption}
\usepackage[
backend=biber,
style=numeric,
sorting=none
]{biblatex}
\usepackage{listings,multicol}  
\usepackage{threeparttable}

\begin{document}

\title{AutoML for Multilayer Perceptron and FPGA Co-design}

\author{\IEEEauthorblockN{1\textsuperscript{st} Philip Colangelo}
	\IEEEauthorblockA{\textit{Intel PSG} \\
		San Jose, USA \\
		philip.colangelo@intel.com}
	\and
	\IEEEauthorblockN{2\textsuperscript{nd} Oren Segal}
	\IEEEauthorblockA{\textit{Hofstra University}\\
		Hempstead, USA \\
		oren.segal@hofstra.edu}
	\and
	\IEEEauthorblockN{3\textsuperscript{rd} Alex Speicher}
	\IEEEauthorblockA{\textit{Hofstra University}\\
		Hempstead, USA \\
		aspeicher1@pride.hofstra.edu}
	\and
	\IEEEauthorblockN{4\textsuperscript{th} Martin Margala}
	\IEEEauthorblockA{\textit{University of Massachusetts Lowell}\\
		Lowell, USA \\
		Martin\_Margala@uml.edu}
}

\maketitle

\begin{abstract}
State-of-the-art Neural Network Architectures (NNAs) are challenging to design and implement efficiently in hardware. In the past couple of years, this has led to an explosion in research and development of automatic Neural Architecture Search (NAS) tools. AutomML tools are now used to achieve state of the art NNA designs and attempt to optimize for hardware usage and design. Much of the recent research in the auto-design of NNAs has focused on convolution networks and image recognition, ignoring the fact that a significant part of the workload in data centers is general-purpose deep neural networks. In this work, we develop and test a general multilayer perceptron (MLP) flow that can take arbitrary datasets as input and automatically produce optimized NNAs and hardware designs. We test the flow on six benchmarks. Our results show we exceed the performance of currently published MLP accuracy results and are competitive with non-MLP based results. We compare general and common GPU architectures with our scalable FPGA design and show we can achieve higher efficiency and higher throughput (outputs per second) for the majority of datasets. Further insights into the design space for both accurate networks and high performing hardware shows the power of co-design by correlating accuracy versus throughput, network size versus accuracy, and scaling to high-performance devices. 
\end{abstract}


\section{Introduction and Motivation}
Optimizing neural network architectures (NNA) is a difficult process in part because of the vast number of hyperparameter combinations that exist. In cases where a combination is not optimal, performance will suffer. Research suggests that the parameters of a network can directly influence the accuracy, throughput, and energy consumption of that model in deployment \cite{canziani2016analysis}. The difficulty in designing performant neural networks has brought a recent surge in interest in the automatic design and optimization of neural networks. The focus of the existing body of research has been on optimizing NNA for accuracy \cite{liu2017hierarchical}\cite{real2017large}\cite{real2018regularized} but lack network-specific hardware optimizations. Our focus is to close this gap by using evolutionary algorithms to search an entire design space, including NNA and reconfigurable hardware. This Co-design approach provides a solution that implements a custom hardware solution for a specific NNA model.

Numerous articles have focused on image classification with convolutional neural networks, in part, due to the introduction of AlexNet in 2012 and the ImageNet database \cite{krizhevsky2012imagenet}. Traditional neural networks such as multilayer perceptrons (MLP) have taken a backseat from the spotlight. Yet, their application is still very relevant, large data-centric companies such as Facebook\cite{hazelwood2018applied}\cite{wu2019machine} and Google \cite{jouppi2018motivation} have published data showing that MLP workloads are the majority of their application base.  Facebook cites the use of MLP for tasks such as determining which ads to display, which stories matter to see in a news feed, and which results to present from a search. Park et al. stress the importance of these networks and the current limitations on standard hardware and the call for what this research aims to solve, i.e., software and hardware co-design in \cite{park2018deep}. Google recently published a performance analysis of their TPU , showing MLP consisting of 61\% of the TPU workloads. Our findings are in line with the results presenting in their paper, stating that MLP workloads are memory bound. Our co-design process is capable of designing optimal hardware structures that balance the compute and memory requirements. Section \ref{section:experiments} provides insights on memory bandwidth considerations.  

At the heart of MLP is a general matrix multiplication (GEMM), and libraries have existed for years such as Basic Linear Algebra Subprograms (BLAS) that provide highly tuned hardware-specific GEMM implementations. Standard hardware such as CPUs and GPUs provide these subroutines to make implementing performant MLPs easy. Still, two problems persist in this typical flow: 1. target hardware is normally not considered during the creation of the MLP, and 2. standard architectures are general purpose. General purpose architectures do not necessarily offer performance scaling concerning the ANN description (see section \ref{section:experiments}). Our research aims to take advantage of the reconfigurable architecture of an FPGA device that is capable of molding to a specific workload and neural network structure. Leveraging evolutionary algorithms to search the entire design space of both MLP and target hardware simultaneously, we find unique solutions that achieve both top accuracy and optimal hardware performance. Section \ref{section:experiments} provides top results for MLP networks and hardware configurations derived from a series of tests over multiple datasets. Correlations between traditional approaches and our approach are discussed to provide insights into how we achieve top results. 

\section{Related Work}
Automating NNA search has been an ongoing effort for the past few decades but is becoming a focus of the NNA research community because of the difficulty in designing deep networks which are ever growing in complexity\cite{real2017large}\cite{real2018regularized}\cite{elsken2018neural}. 
Automatic Artificial Neural Network Architectures Search (NAS) can be conducted using different strategies such as random search, evolutionary algorithms, Reinforcement Learning (RL), Bayesian optimization, 
and gradient-based methods \cite{elsken2018neural}.
Using Evolutionary Algorithms (EAs) to search for performant architectures has been investigated extensively \cite{miller1989designing}\cite{stanley2002evolving} over the years. 
Some recent results indicate that evolutionary algorithms offer better results than random search and reinforcement learning \cite{real2018regularized}.
Recently, there has been growing interest in NAS for deep neural networks that specialize in image recognition \cite{liu2017hierarchical}\cite{real2017large}\cite{Zoph2018LearningTA}. A recent survey on available tool flows is available here \cite{venieris2018toolflows}. The body of work on NAS concentrate on accuracy as the main measure of performance, though optimizing for NAS can lead to more simplified NNA that could in turn simplify and optimize hardware designs \cite{real2018regularized}\cite{elsken2018neural}. On the other hand, optimizing for hardware performance parameters (latency/throughput/power) is normally done on an existing NNA design and there is no attempt to modify the NNA (layers/neurons etc.)\cite{venieris2018toolflows}.   

\section{Evolutionary Cell Aided Design Flow}
\begin{figure}
	\center
	\includegraphics[scale=0.42]{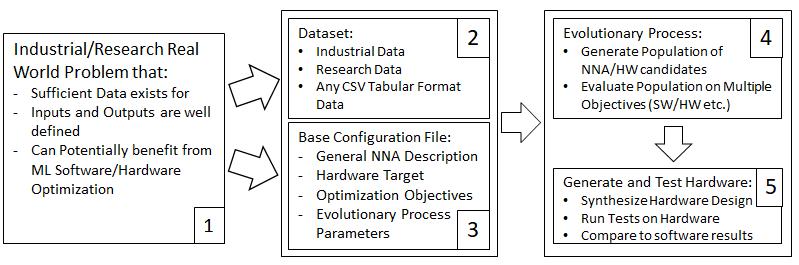} 						
	\caption{ECAD flow.}
	\label{fig:ecad_flow}	
\end{figure}

The Evolutionary Cell Aided Design (ECAD) flow, previously described in \cite{Colangelo:2019:HPEC}, is shown in Figure \ref{fig:ecad_flow} and starts with a general industrial/research problem that (a) sufficient data exists for (b) there are well defined inputs/outputs and (c) it is a problem that can benefit from software/hardware acceleration. Once such a problem is identified a dataset will be exported into a Comma Separated Value (CSV) tabular data format, in addition a configuration file will be created and will contain information on (a) the general NNA structure including input and output sizes, initial number of layers and neurons, (b) Hardware target including reconfigurable hardware device type, DSP count, memory size and number of blocks, (c) optmization targets such as accuracy, throughput, latency, and floating-point operations. Note that the configuration file can be generated automatically based on an existing template configuration file and the dataset.

\subsection{Evolutionary Process} The ECAD Evolutionary process, based on a steady-state model \cite{goldberg1991comparative}, generates a population of NNA/Hardware co-design candidates each with a complete set of parameters that effect both the accuracy and the hardware performance. The parameters we considered during our searches included number of layers, layer size, activation function, and bias. Each candidate in the population is evaluated according to configurable and potentially multiple criteria, for example accuracy alone or accuracy vs throughput. The \textit{raw} evaluation measurments are done on a software artifact, dubbed a  \textit{Worker} in ECAD terminology. The Worker returns the \textit{raw} evaluation information to a \textit{Master} process. The \textit{Master} process orchestrates the evaluation process by distributing the co-design population and by evaluating the results. Result evaluation is done using user defined fitness functions. For example, an accuracy fitness function can simply return the accuracy value obtain by a simulation worker(see next section for more information on worker types). But it can also scale or weight the value or specify to minimize or maximize the value. Simple evaluations functions can be specified in the configuration file and more complex ones are written in code and added by registering them with the framework.



\subsection{ECAD Hardware} \label{section:HW}

The evolutionary search has three workers at its disposal to assess the fitness of various hardware platforms. The simulation worker is useful for assessing instruction-set based architectures such as CPU and GPU, whereas the physical and hardware database workers are useful for hardware that requires design and synthesis.

Simulation workers take queries from the evolutionary engine and run them on the target hardware. For example, if the target hardware is a GPU, the evolutionary engine sends the physical worker an ANN description who converts it into a readable format for the GPU. Once the GPU returns the result, the physical worker is responsible for recording all necessary metrics that are provided back to the engine to assess the fitness. Metrics include throughput, latency, and power consumption. 

Hardware database workers provide a means for hardware platforms that are easily simulated or modeled. In our experiments, for example, we leveraged the hardware database worker to provide a means of accepting both an ANN description and hardware configuration that together were run through a model to obtain the metrics for fitness evaluation. Once the engine finds performant designs via tracking each configuration's fitness, the model can be realized into hardware and run through the physical worker. FPGA proved to be a suitable architecture for the hardware database worker. The reconfigurable nature of FPGAs, coupled with modeled overlay architectures, allows the worker to assess many configurations in a relatively swift manner compared to running through synthesis tools. 

Physical workers can be used to synthesize and evaluate hardware designs. While the hardware database worker provides fitness of the overall application with metrics such as throughput, the physical worker aims to provide the fitness of the hardware design itself through metrics such as power, logic utilization, and operation frequency. In the case of Intel FPGAs, the physical worker responds with ALM, M20K, and DSP utilization, power estimations, and clock frequency (Fmax).  

Having access to these workers in the evolutionary engine is a powerful tool that provides various performance metrics about the hardware platform as a function of the ANN. Since ANN descriptions can vary widely, the throughput and power consumption on hardware devices can also vary widely. Understanding how a hardware device performs across ANN variations aids the decision on how to deploy ANN solutions, as seen in the experiments. 

Beyond these tools providing insights into architecture performance, the Pareto frontiers that result after parsing the evolutionary design space define what the optimal solution is. Every industry deploying ANN solutions may have different opinions and demands for their use case. One size fits all, or one solution is not realistic or optimal. Having the data to make decisions based on trade-offs is highly valuable. Since some ANN descriptions are approximations, i.e., they include redundancy (although we note that newer research is starting to limit the amount of redundancy) and educated guesses, it is an unreasonable expectation to hand optimize ANN descriptions for a specific hardware platform. 

\subsection{FPGA Design Model and Hardware Description} \label{section:model}

Workers can provide input to any hardware capable of being modeled efficiently. Given the reconfigurable architecture of FPGAs and modern HLS tools capable of describing overlay designs that interface nicely with software programs, we chose Intel FPGAs and OpenCL as our hardware platform and development. Any flavor of FPGA can use the overlay-style architecture we developed, and all that is required to change the design search space, for example, between Arria 10 and Stratix 10, is the hardware configuration used by the hardware database worker. 

Hardware database workers receive configurations that include enough information to construct the model and derive the necessary performance results. Part of this configuration is a hardware-specific configuration file that defines the target accelerator. This information includes the name of the FPGA, the relevant primitive logic details such as DSP and SRAM count, target clock frequency, the type of global memory (DRAM) to be used, and its speed and rate. All of this information is used to estimate performance.  The other part of the configuration is the ANN description. Low-level hardware suitable parts are decomposed from the ANN description and define the work that must be done by the accelerator. 

General matrix multiplications are commonly used to accelerate multilayer perceptrons, so the design we used is based on a 2D systolic array architecture that includes additional functionality to support activation functions and vector additions for bias operations. This ``grid" architecture has various design space variables that we allow mutations to take place on. The variables are the number of rows and columns, double buffer cache sizes for each dimension, called interleaving, and the vector width of each processing element (PE). Every variable has implications in the design space. Each mutation has a unique fitness, not only as it applies to the hardware but also the ANN. This works leverages and build on the SGEMM architecture described in \cite{vishwanath2016enabling}.

Our model returns values we deemed fundamental, including potential and effective performance,  total time, outputs per second, and latency. Giga-operations per second (GOP/s) is the unit to describe the potential and effective performance. The potential performance is typically the marketed performance that defines the roofline of the configuration, whereas the effective performance defines the actual or real performance of the configuration under a workload. The delta between them provides a metric on how well the problem maps to the model. The total time for our model is defined by the difference between the timestamp when all necessary data persists in the accelerator DRAM, and we enqueue the OpenCL kernels, and the timestamp once all results persist in DRAM and the last OpenCL kernel returns. We call this a run. Outputs per second is a generalization of the data type and provides the total results we can achieve in one second. For example, if the data type is images, then this metric could be interpreted as images per second. Lastly, latency is the time it takes from the start of a run to store one result into DRAM.  

Calculating these results in the model is accomplished by starting with the baseline performance of a configuration. All data is 32-bit floating-point and maps into the hardened floating-point DSP blocks of the Arria 10 device. From \cite{vishwanath2016enabling}, we can calculate the baseline performance by determining how many DSP blocks are doing work. The utilization of DSPs is the product of the grid dimensions and vector width.  This number is the potential performance, but before considering bandwidth. Using the DRAM specs from the configuration, we can determine the ratio of how much bandwidth is available to how much we need. Cycles per block of data divided into the size of a block in bytes are used to calculate bandwidth needs. This calculation is the potential performance. Next, the grid configuration is used to break the ANN up into a series of blocked matrix multiplications. Having the data now blocked, we can derive all other results. 

\subsection{MLP Mapping to Hardware} 

GEMM nomenclature can be used to describe the three key dimensions that make up the problem size for MLP layers. Two matrices A and B are multiplied together to form a new matrix C. If matrix A has sizes \textit{m x k} and matrix B \textit{k x n}, then matrix C is \textit{m x n}. \textit{M} is the number of inputs that are processed at once and commonly referred to as the batch size. Architectures such as GPU typically batch with a larger \textit{M} dimension to fill up compute cores and obtain higher throughput. Our design for FPGA does not need to increase batching because the PEs can be arranged in a manner that exploits parallelism in other dimensions. This results in a lower batch and lower latency accelerator. With sufficient bandwidth, hardware efficiency can be increased by keeping the \textit{m} dimension smaller, and further vectorizing on \textit{k} and \textit{n} dimensions. \textit{N} dimension is the number of neurons that also defines a subsequent layer \textit{k}. Lastly, the size of the dataset defines the first layer \textit{k}. Fitting an MLP into hardware is one aspect that makes the co-design approach so powerful.

\section{Experiments} \label{section:experiments}

In this section, we show the results from running a series of evolutionary searches on six different data sets: MNIST \cite{dataset:mnist}, Fashion MNIST \cite{dataset:fmnist}, Credit-g \cite{dataset:creditg_phishing}, Har \cite{dataset:har}, Phishing \cite{dataset:creditg_phishing}, and Bioresponse \cite{dataset:bioresponse}. MNIST and Fashion MNIST were chosen so we could compare against wildly used research benchmarks. Credit-g/Har/Phishing/Bioresponse were chosen since they represent potential real-world datasets.  

FPGA hardware results assume the architecture described in section \ref{section:HW}. Search was done on two different FPGAs, an Arria 10 1150 GX device \cite{Arria10} at a clock frequency of 250 MHz and a Stratix 10 2800 device at a clock frequency of 400 MHz. After many hardware compiles, 250 MHz was, on average, the frequency the OpenCL design achieved for the Arria 10. Running at 250 MHz provides a peak throughput of 759 GFLOP/s FP32 single-precision performance. The accelerator card was the development kit from Intel, which contains a single bank of DDR4 memory, providing a peak bandwidth of 19.2 GB/s. In many cases, the evolutionary algorithm requested configurations that led to bandwidth-constrained designs. We do provide a few results that included a design space with 2 and 4 DDR banks providing 38.4 and 76.8 GB/s bandwidth respectively. All Stratix 10 models were run with 4 banks of DDR. 

GPU results were obtained using three different devices. An NVIDIA Quadro M5000 with 8GB DDR5 capable of 4.3 TFLOP/s of FP32 single-precision performance with 211 GB/s memory bandwidth, a Titan X capable of 12 TFLOP/s of FP32 single-precision performance and Radeon VII 16GB HBM2, 13.44 FP32 TFLOPS and 1 TB/s memory bandwidth. Profiling for GPU was done using trace files generated from Tensor Flow. The timing report considers matrix multiplication, activation, and vector addition routines, but it does not appear to take into account DRAM transfers. FPGA timing reports do consider DRAM because memory buffering is an active component in the design. Overall, conclusions are not affected by the difference between timing reports that likely skews direct comparisons of FPGA and GPU (in favor of GPU). 

Power consumption for GPU was captured during runs using the built-in nvidia-smi utility. We found the power management for GPU to be efficient since, in most cases, the effective performance was rather low (see section \ref{section:eff}, and so was the power. On average, the GPU power measured around 50 W for the 150 W (maximum) device. Quartus' Power Analyzer Tool was used to capture power estimates for FPGA. Across the many Arria 10 designs compiled, we found the minimum power to be 22.5 W, maximum power to be 31.89 W, and average power to be 27 W. The difference between power numbers for FPGA and GPU was that FPGA represents chip power and GPU represents total board power. This discrepancy made it challenging to compare power between architectures. Given this discrepancy and the small variance across configurations, we decide to leave the topic of power outside the conclusions. 

\subsection{Overall Performance Results}
We first examine the accuracy results we were able to achieve using the evolutionary flow. Table \ref{table:topacc10fold} present the top results obtained from the evolutionary algorithm searching for accuracy using a 10-fold evaluation method\cite{vanschoren}\cite{kohavi1995study}. This method splits the data set into 10-equal train/test folds and measures performance on each of the folds. As can be seen, ECAD MLP (our method) was able to achieve better accuracy results then all MLP based classifiers. In addition, the ECAD evolutionary process managed to outperform non-MLP based methods for the credit-g and phishing datasets.
The mnist and fashion-mnist datasets were obtained outside of openml use a traditional 1-fold train/test dataset and we compare them to results published in the literature\cite{dataset:fmnist}\cite{dataset:mnist}. As can be seen, our mnist and fashion-mnist accuracy results outperform the top reported results. In addition our auto MLP network has the second best reported result and is 0.0047 shy of the SVC method record holder.
Table \ref{table:runtime} shows ECAD run time statistics for the results reported in Tables \ref{table:topacc10fold} and \ref{table:topacc1fold}. It reports the number of different NNA/HW combinations that were automatically generated and evaluated by the ECAD system, the average time per evaluation and total evaluation time of all candidate architectures. Note that in order to optimize the search and run time of the system, potential NNA/HW candidates are first analyzed for similarities to previous evaluations and duplicates are not evaluated twice.

Table \ref{table:pareto} shows the results for two top Pareto frontier solutions for each data set. Note that this search was done outside of the OpenML spec, i.e., running a single fold. The solutions provide accuracy and throughput for a Stratix 10 (S10) FPGA and TitanX (TX) GPU. In the majority of cases the FPGA achieved higher performance than the GPU. Credit-g, for example, favored GPU for higher accuracy, but looking at the second row for credit-g, by sacrificing just one point of accuracy, the FPGA sees a very significant improvement in throughput. 

\bgroup
\def\arraystretch{1.2}
\begin{table*}[htb]
\centering
\caption{Top 10-fold Accuracy (Acc) for All Datasets Compared to Previous Works}
\label{table:topacc10fold}
\begin{tabular}{|c|c|c|c|c|c|c|c|}
\hline
\textbf{Dataset} & \textbf{Top Acc (Any)} & \textbf{Top Method } & \textbf{Top Acc (MLP)}   & \textbf{MLP Type} & \textbf{ECAD MLP }  \\ \hline
Credit-g      & 0.7860               				& mlr.classif.ranger				& 0.7470                & *MLPClassifier           & \textbf{0.7880}    \\ \hline
Har           & 0.9957               				& *DecisionTreeClassifier	        & 0.1888                & *MLPClassifier           & 0.9909             \\ \hline
Phishing      & 0.9753               				& *SVC 						        & 0.9733                & *MLPClassifier           & \textbf{0.9756}    \\ \hline
Bioresponse   & 0.8160               				& mlr.classif.ranger				& 0.5423                & *MLPClassifier           & 0.8038            \\ \hline
\end{tabular}
 \begin{tablenotes}
	\small
	\item \textit{Note} The OpenML datasets/results can be found at openml.org: credit-g(https://www.openml.org/t/31), har(https://www.openml.org/t/14970), Phishing(https://www.openml.org/t/34537) and Bioresponse(https://www.openml.org/t/14966). Entries with * denote models from sklearn.
 \end{tablenotes}
\end{table*}
\begin{table*}[htb]
	\centering
	\caption{Top 1-fold Accuracy (Acc) for All Datasets Compared to Previous Works}
	\label{table:topacc1fold}
	\begin{tabular}{|c|c|c|c|c|c|c|c|}
		\hline
		\textbf{Dataset} & \textbf{Top Acc (Any)} & \textbf{Top Method } & \textbf{Top Acc (MLP)}   & \textbf{MLP Type} & \textbf{ECAD MLP} \\ \hline
		MNIST         & 0.9979             				    & Manual  							& 0.9840                & Manual(no distortions)   & 0.9852 \\ \hline
		Fashion MNIST & 0.8970               				& SVC  								& 0.8770                & MLPClassifier            & 0.8923 \\ \hline
	\end{tabular}
	\begin{tablenotes}
		\small
		\item \textit{Note} MNIST and Fashion MNIST are standalone pre-split(1-fold) datasets obtained from Keras dataset collection (https://www.tensorflow.org/api\_docs/python/tf/keras/datasets).
	\end{tablenotes}
\end{table*}

\egroup

\bgroup
\def\arraystretch{1.2}
\begin{table*}[htb]
	\centering
	\caption{Top Accuracy Run Time Statistics}
	\label{table:runtime}
	\begin{tabular}{|c|c|c|c|c|c|}
		\hline
		\textbf{Dataset} & \textbf{Total Models Evaluated } & \textbf{AVG Model Evaluation Time (s)}   & \textbf{Total Evaluation Time (s)} \\ \hline
		MNIST         & 553  			& 71.23                     & 39388.6 \\ \hline
		Fashion MNIST & 481  			& 82.55                     & 39708.7 \\ \hline
		Credit-g      & 10480			& 2.24                      & 23495.2 \\ \hline
		Har           & 3229	        & 10.20                     & 33069.4  \\ \hline
		Phishing      & 3534            & 9.24                      & 32661.3  \\ \hline
		Bioresponse   & 5309            & 5.89                      & 31285.0  \\ \hline
	\end{tabular}
	\begin{tablenotes}
		\small
		\item \textit{Note} Each model generated is a fully functional combination of NNA traits and hardware traits that is evaluated for performance on any of the measured metrics. The ECAD system \textit{caches} similar configurations and avoids reevaluating them.
	\end{tablenotes}
\end{table*}
\egroup

\bgroup
\begin{table}[htb]
\centering
\caption{Best Pareto Frontier Results for Searching Accuracy and Throughput}
\label{table:pareto}
\begin{tabular}{|c|c|c|c|}
\hline
\textbf{Dataset}       & \textbf{Accuracy} & \textbf{S10 (output/s)} & \textbf{TX (output/s)}  \\ \hline
MNIST         & 0.9841            & 7.97E5                         & 7.73E5                   \\ \hline
MNIST         & 0.9763            & 2.45E6                         & 1.97E6                   \\ \hline
Fashion MNIST & 0.893               & 4.8E5                            & 8.1E5                   \\ \hline
Fashion MNIST & 0.8850               & 1.92E6                            & 2.3E6                   \\ \hline
Har           & 0.996             & 1.16E6                         & 9.59E5                   \\ \hline
Har           & 0.985             & 4.74E6                         & 2.46E6                   \\ \hline
Credit-g      & 0.83              & 8.19E3                         & 1.59E6                 \\ \hline
Credit-g      & 0.82              & 1.40E7                         & 1.23E6                   \\ \hline
Bioresponse   & 0.798             & 4.64E5                         & 1.34E6                   \\ \hline
Bioresponse   & 0.7952            & 1.36E6                         & 1.66E6                   \\ \hline
Phishing      & 0.9675            & 6.81E6                         & 2.27E6               \\ \hline
Phishing      & 0.9656            & 1.16E7                         & 2.27E6                \\ \hline
\end{tabular}
\end{table}
\egroup

\subsection{Accuracy versus Throughput}
\begin{figure}
\centering
\begin{subfigure}[tb]{\columnwidth}
	\includegraphics[width=\columnwidth]{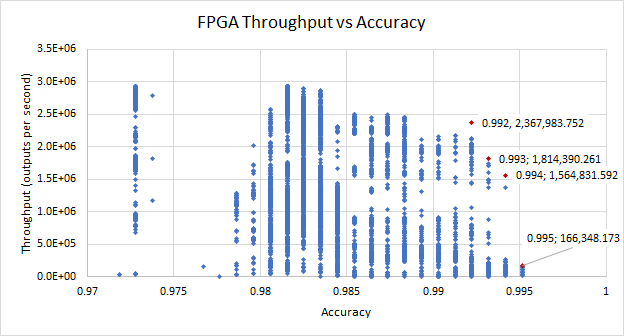} 	
	\caption{}
	\label{fig:har_fpga_perf_acc}	
\end{subfigure}
\begin{subfigure}[tb]{\columnwidth}
	\includegraphics[width=\columnwidth]{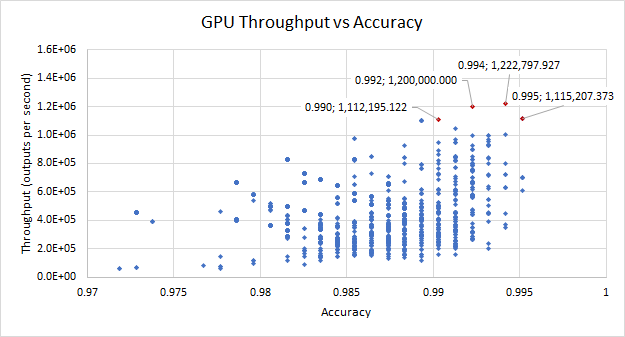} 	
	\caption{}
	\label{fig:har_gpu_perf_acc}	
\end{subfigure}
\caption{Performance of FPGA and GPU at different levels of accuracy for the har dataset}
\end{figure}

Part of the motivation for this research is to show the adaptability of reconfigurable hardware to a neural network. This flexibility and molding over a series of GEMM calls produce particular results aimed to fit an optimal network description, i.e., highest accuracy with an optimal hardware solution producing high and efficient throughput. We begin running the evolutionary search over the HAR dataset and observe how both GPU and FPGA reacted to each evolutionary step. Figure \ref{fig:har_fpga_perf_acc} provides results for Arria 10 and Figure \ref{fig:har_gpu_perf_acc} provides the results for the Quadro M5000. 

The evolutionary process provided many high performance results in terms of accuracy with the top at 0.995 and provided many varying results for throughput. The broad swing of throughput results (especially for GPU, being a fixed architecture) shows that many MLP solutions exist that achieve the top accuracy. GPUs accelerate each solution in the same way, so the varying levels of throughput mean the MLP structure is changing. The top throughput GPU achieved at an accuracy of 0.995 is shown to be approximately 1E6 results per second, while the bottom throughput result is approximately 6E5 results per second. For the same reasoning, as we move down a point of accuracy, the GPUs performance hardly changes, this is because of the number of neurons remains roughly the same, it is the distribution between layers causing the effect on accuracy. For GPU, there is roughly no relationship between the number of neurons and the throughput. FPGA has a different correlation. Figure \ref{fig:har_fpga_perf_acc} shows that the distribution of neurons in an MLP greatly affects the performance. Unlike GPU, there is (potentially) a different hardware configuration for each data point. While top accuracy only reaches 1.6E5 outputs per second, moving down accuracy just 0.1\% results in a giant leap to 1.56E6 outputs per second, an order of magnitude higher. Further, moving down another 0.2\% results in approximately another 1.5x in performance. Each dataset tested displayed this same trend. 

\subsection{Performance Scaling against Bandwidth}
\begin{figure}[tb]
	\center
	\includegraphics[width=\columnwidth]{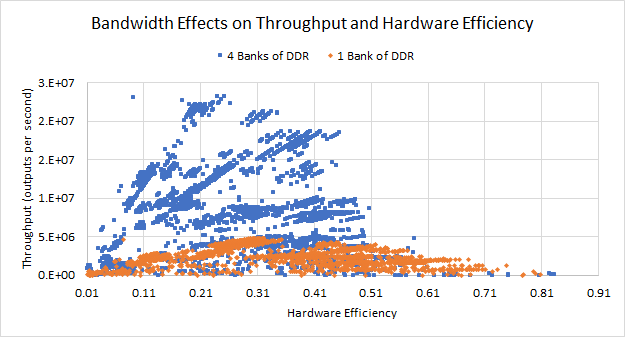}		
	\caption{Throughput and hardware efficiency for FPGA designs with 1 and 4 banks of DDR on the credit-g data set}
	\label{fig:bw_effect_perf}	
\end{figure}

Most designs returned by the evolutionary algorithm tended to be smaller, using only a percentage of the available DSPs. The reason is that scaling to more DSPs requires more data, which requires more memory bandwidth. We hit the memory bandwidth roofline many times due to only having a single bank of DDR. Observing this, we ran a series of tests that provided the hardware database model with 2 and 4 banks of DDR. We found mostly a linear scaling going from 1 to 4, so we show the effect bandwidth has on throughput for 1 and 4 DDR banks in Figure \ref{fig:bw_effect_perf}. Higher bandwidth did not produce greater efficiency but did result in higher throughput overall. 

\subsection{Hardware Efficiency and Scaling to Larger Devices} \label{section:eff}
We modified the hardware database worker to return results based on a Stratix 10 2800 device with 4 banks of DDR. This device offers up to a 10x performance scaling from the Arria 10 device we used in previous experiments. While the Stratix 10 device is capable of performing up to 10 TFLOP/s, we used the same methodology as the Arria 10 device to search Stratix 10 devices at a clock frequency of 400 MHz scaling back the roofline to  4.6 available TFLOP/s. To better match the performance of the Stratix 10 device, we used an Nvidia Titan X device capable of 12 TFLOP/s. 

We found that overall, the reconfigurable architecture had a more significant potential to perform at a higher level; however, for some data sets, the Titan X raw throughput was able to achieve the highest outputs per second. Throughput is a popular metric to consider, but we also found efficiency to be of high importance. The reason the larger FPGA device handled running these datasets through an MLP is due to the allocation of resources. When the evolutionary algorithm chooses a hardware configuration, there is an allotted space on the FPGA that contains a potential performance. Then after mapping the MLP, we get an effective performance (see section \ref{section:model} for details on potential and effective performance). The ratio of effective performance over potential performance gives us hardware efficiency. Efficiency could yield fewer FPGA resources while maintaining throughput leaving logic available for other tasks such as pre or post-processing. 

\begin{figure}[tb]
	\center
	\includegraphics[width=\columnwidth]{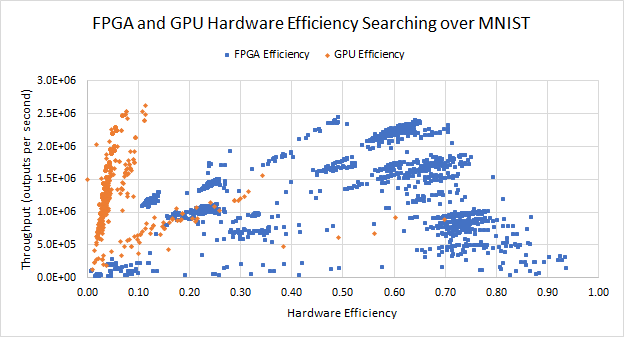}	
	\caption{Hardware efficiency results for a Stratix 10 2800 and Titan X searching over the MNIST dataset}
	\label{fig:fpga_gpu_eff_mnist}	
\end{figure}

Figure \ref{fig:fpga_gpu_eff_mnist} shows the FPGA and GPU efficiency of various solutions from searching across the MNIST dataset. The top accuracy for this particular run was 0.9845, and  throughput was 796,611 and 773,162 outputs per second for the FPGA and GPU respectively, almost identical. If we consider efficiency for this result, the FPGA utilized 41.5\% of the allocated logic, while the GPU only utilized 0.3\%. We calculated GPU efficiency as the number of operations per second obtained from a run out of the total potential operations per second of the device. The conclusion is that without target hardware in mind during MLP development, there is a good chance of losing efficiency. Running an evolutionary search on reconfigurable hardware can balance the fitness between throughput and accuracy. 

\section{Conclusions}

We address the difficulty of designing highly performant neural networks by leveraging evolutionary search algorithms capable of finding the fittest solutions for both classification accuracy and hardware throughput. This process is shown to be both highly efficient and effective compared to traditional approaches that first design a neural network to achieve a target accuracy, then run it on general-purpose hardware. Through a series of experiments, we present our results for state of the art neural network configurations that surpass current published work. We explain the power of co-design by discussing the results of experiments showing accuracy versus throughput, performance scaling with bandwidth, and scaling designs with larger devices. 

\section{Future Directions}

We believe a custom co-design framework for MLPs and reconfigurable hardware could be of use to a large audience. We plan to continue and develop this framework, streamline its use and make it available to the general public.

\printbibliography
\end{document}